%% file: aaai_submission.tex
\pdfoutput=1

\def\year{2019}\relax
\documentclass[letterpaper]{article} 
\usepackage{aaai19}  
\usepackage{times}  
\usepackage{helvet}  
\usepackage{courier}  
\usepackage{url}  
\usepackage{graphicx}  
\usepackage{amsmath,amssymb} 
\usepackage{algorithm}
\usepackage{algorithmic}

\graphicspath{{figs/}}

\begin{document}
%
\title{Improving GAN with neighbors embedding and gradient matching}
\author{Ngoc-Trung Tran$^*$, Tuan-Anh Bui\thanks{These authors have contributed equally to this work}, Ngai-Man Cheung\\
ST Electronics - SUTD Cyber Security Laboratory\\
Singapore University of Technology and Design
}
\maketitle

\input{abstract}
\input{introduction}
\input{related_works}
\input{method}
\input{experiment}
\input{conclusion}

\input{acknowledge}

\bibliographystyle{aaai}
\bibliography{../../../biblio/biblio}

\end{document}

%% file: abstract.tex
\begin{abstract}


We propose two new techniques for training Generative Adversarial Networks (GANs). Our objectives are to alleviate mode collapse in GAN and improve the quality of the generated samples. First, we propose
{\em neighbor embedding}, a manifold learning-based regularization to explicitly retain local structures of latent samples in the generated samples. This prevents generator from producing nearly identical data samples from different latent samples, and reduces mode collapse. 
We propose an 
{\em inverse} t-SNE regularizer to achieve this.
Second, we propose a new technique, {\em gradient matching}, to align the distributions of the generated samples and the real samples. As it is challenging to work with high-dimensional sample distributions, we propose to align these distributions through the scalar discriminator scores. We constrain the difference between the discriminator scores of the real samples and generated ones. We further constrain the difference between the gradients of these discriminator scores. We derive these constraints from Taylor approximations of the discriminator function.
We perform experiments to demonstrate that our proposed techniques are computationally simple and easy to be incorporated in existing systems.
When Gradient matching and Neighbour embedding are applied together, our GN-GAN achieves outstanding results on 1D/2D synthetic, CIFAR-10 and STL-10 datasets, e.g. FID score of $30.80$ for the STL-10 dataset.
Our code is available at: https://github.com/tntrung/gan



\end{abstract}

%% file: introduction.tex
\section{Introduction}

Generative Adversarial Networks (GANs) \cite{goodfellow-nisp-2014,goodfellow-nips-2016} are  popular methods for training generative models. GAN training is a two-player minimax game between the discriminator and the generator. While the discriminator learns to distinguish between the real and generated (fake) samples, the generator creates samples to confuse the discriminator to accept its samples as ``real". This is an attractive approach. However,  stabilizing the training of GAN is still an on-going important research problem.

Mode collapse is one of the most challenging issues when training GANs. Many advanced GANs have been proposed to improve the stability \cite{nowozin-nips-2016,arjovsky-arxiv-2017,gulrajani-arxiv-2017}. However,  mode collapse is still an issue.

In this work, we propose two techniques to improve GAN training. 
First, inspired by t-distribution stochastic neighbors embedding (t-SNE) \cite{maaten-jmlr-2008},
which is a well-known dimensionality reduction method, we propose an {\em inverse} t-SNE regularizer to reduce mode collapse.
Specifically,  while t-SNE aims to preserve the  structure of the high-dimensional data samples in the reduced-dimensional manifold of latent samples, 
we reverse the procedure of t-SNE to 
explicitly retain local structures of latent samples in the high-dimensional generated samples. This prevents generator from producing nearly identical data samples from different latent samples, and reduces mode collapse. 
Second, we propose a new objective function for the generator by aligning the real and generated sample distributions, in order to generate realistic samples. We achieve the alignment via minimizing the difference between the discriminator scores of the real samples and generated ones. 
By using the discriminator and its scores, we can avoid working with high-dimensional data distribution. We further constrain the difference between the gradients of discriminator scores. We derive these constraints from Taylor approximation of the discriminator function. Our principled approach is significantly different from the standard GAN \cite{goodfellow-nisp-2014}: our generator does not attempt to directly fool the discriminator; instead, our generator produces fake samples that have similar discriminator scores as the real samples. 
We found that with this technique the distribution of the generated samples approximates well that of the real samples, and the  generator can produce more realistic samples.




%% file: related_works.tex
\section{Related Works}

Addressing issues of GANs \cite{goodfellow-nips-2016}, including gradient vanishing and mode collapse, is an important research topic. A  popular direction is to focus on improving the discriminator objective. The discriminator can be formed via the f-divergence \cite{nowozin-nips-2016}, or distance metrics \cite{arjovsky-arxiv-2017a,bellemare-arxiv-2017}.  And the generator is trained by fooling the discriminator via the zero-sum game. Many methods in this direction have to regularize their discriminators; otherwise, they would cause instability issues, as the discriminator often converges much faster than the generator. Some regularization techniques are weight-clipping \cite{arjovsky-arxiv-2017a}, gradient penalty constraints \cite{gulrajani-arxiv-2017,roth-nips-2017,kodali-arxiv-2017,petzka-arxiv-2017,liu-arxiv-2018}, consensus constraint, \cite{mescheder-nips-2017,mescheder-icml-2018}, or spectral norm \cite{miyato-iclr-2018}. However,  over-constraint of the discriminator may cause the cycling issues \cite{nagarajan-nips-2017,mescheder-icml-2018}.

Issues of GAN can also be tackled via the optimizer regularization: changing optimization process \cite{metz-arxiv-2016}, using two-time scale update rules for better convergence \cite{heusel-arxiv-2017}, or averaging network parameters \cite{yazici-arxiv-2018}.

Regularizing the generator is another direction: i) It can be achieved by modifying the generator objective function with feature matching \cite{salimans-nisp-2016} or discriminator-score distance \cite{tran-eccv-2018} ii) Or,  using Auto-Encoders (AE) or latent codes to regularize the generator. AAE \cite{makhzani-arxiv-2015} uses AE to constrain the generator. The goal is to match the encoded latent distribution to some given prior distribution by the minimax game. The problem of AAE is that pixel-wise reconstruction with $\ell_2$-norm would cause the blurry issue. And the minimax game on the latent samples has the same problems (e.g., mode collapse) as on the data samples. It is because  AE alone is not powerful enough to overcome these issues. VAE/GAN \cite{larsen-arxiv-2015} combined VAE and GAN into one single model and used feature-wise distance for the reconstruction to avoid the blur. The generator is regularized in the VAE model to reduce the mode collapse. Nevertheless, VAE/GAN has the similar limitation of VAE \cite{kingma-arxiv-2013}, including re-parameterization tricks for back-propagation, or, requirement to  access to an exact functional form of prior distribution. ALI \cite{dumoulin-arxiv-2016} and BiGAN \cite{donahue-arxiv-2016} jointly train the data/latent samples in GAN framework. This method can learn the AE model implicitly after training. MDGAN \cite{che-arxiv-2016} required two discriminators for two separate steps: manifold and diffusion. The manifold step manages to learn a good AE. The diffusion step is similar to the original GAN, except that the constructed samples are used as real samples instead. InfoGAN \cite{chen-arxiv-2016} learned the disentangled representation by maximizing the mutual information for inducing latent codes. MMGAN \cite{park-icpr-2018} makes strong assumption that manifolds of real and fake samples are spheres. First, it aligns real and fake sample statistics by matching the two manifold spheres (centre and radius), and then it applies correlation matrix to reduce mode collapse. Dist-GAN \cite{tran-eccv-2018} constrains the generator by the regularized auto-encoder. Furthermore, the authors use the reconstructed samples to regularize the convergence of the discriminator. 

Auto-encoder can be also used in the discriminator objectives. EBGAN \cite{zhao-arxiv-2016} introduces the energy-based model, in which the discriminator is considered as the energy function minimized via reconstruction errors. BEGAN \cite{berthelot-arxiv-2017} extends EBGAN by optimizing Wasserstein distance between AE loss distributions.


%% file: method.tex
\section{Proposed method}

Our proposed system with gradient matching (GM) and neighbor embedding (NE) constraints, namely GN-GAN, 
consists of three main components: the auto-encoder, the discriminator, and the generator. In our model, we first train the auto-encoder, then the discriminator and finally the generator as presented in Algorithm \ref{alg-01}. 

\begin{algorithm}
 \footnotesize
 \caption{Our GN-GAN model}
 \begin{algorithmic}[1]
 \STATE Initialize discriminator, encoder and generator $D, E, G$ respectively. $N_{iter}$ is the number of iterations.
 \REPEAT
  \STATE $\mathrm{x} \leftarrow$ Random mini-batch of $m$ data points from dataset.
  \STATE $\mathrm{z} \leftarrow$ Random $n$ samples from noise distribution $P_\mathrm{z}$
  \STATE // \textit{Training the auto-encoder using $\mathrm{x}$ and $\mathrm{z}$ by Eqn. \ref{eq-regularized-autoencoder}}
  \STATE $E, G \leftarrow \min \mathcal{V}_{AE}(E,G)$
  \STATE // \textit{Training discriminator according to Eqn. \ref{eq_D_log_obj} on $\mathrm{x}, \mathrm{z}$} 
  \STATE $D \leftarrow \max \mathcal{V}_D(D,G)$  
  \STATE // \textit{Training the generator on $\mathrm{x}, \mathrm{z}$  according to Eqn. \ref{eq:genobj_new}.} 
  \STATE $G \leftarrow \min \mathcal{V}_G(D,G)$  
 \UNTIL $N_{iter}$
 \RETURN $D, E, G$ 
 \end{algorithmic} 
 \label{alg-01}
\end{algorithm}

\subsection{Neighbors embedding constraint for Auto-encoder}

We use auto-encoder (AE) in our model for two reasons: i) to prevent the generator from being severely collapsed. ii) to regularize the generator in producing samples that resemble real ones. However, using AE alone is not adequate to avoid mode collapse, especially for high-dimensional data. Therefore, we propose additional regularization as in Eq. \ref{eq-regularized-autoencoder}:

\begin{equation}
V_{AE}(E,G) = ||\mathrm{x} - G(E(\mathrm{x}))||^2 + \lambda_\mathrm{r} V_R(E,G)
\label{eq-regularized-autoencoder}
\end{equation}

Eq. \ref{eq-regularized-autoencoder} is the 
objective of our regularized AE. The first term is  reconstruction error in conventional AE.
The second term $V_R(E,G)$ is our proposed neighbors embedding constraint, to be discussed.  Here, $G$ is GAN  generator (decoder in AE), $E$ is the encoder and $\lambda_\mathrm{r}$ is a constant.

\begin{figure}
\centering
\includegraphics[scale=0.5]{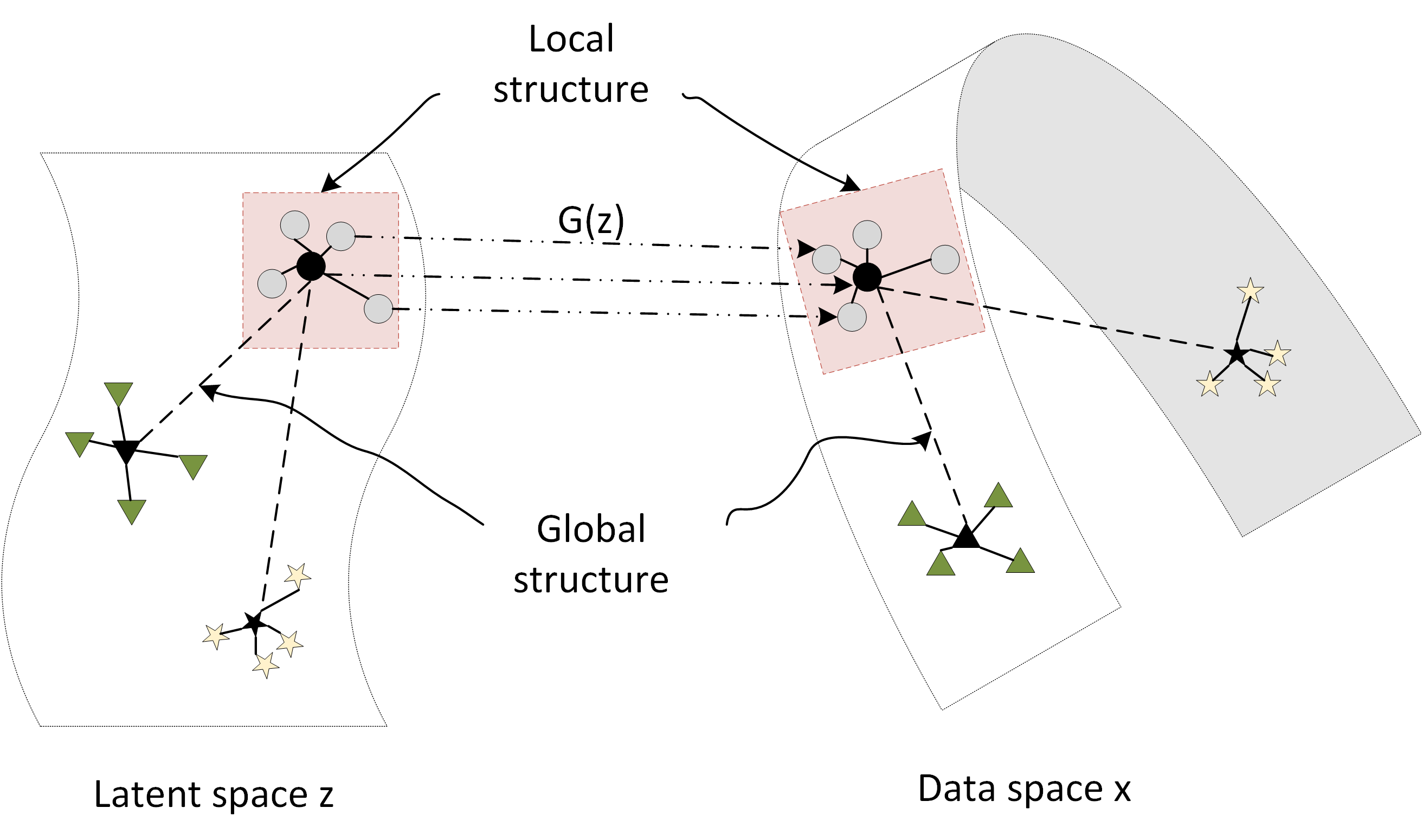}
\caption{Illustration of the neighbor-embedding (NE) constraint. NE  regularizes the generator to produce high-dimensional data samples such that latent sample distance and data sample distance are consistent.}
\label{neighbor_embedding}
\end{figure}


Mode collapse is a failure case of GAN when the generator often generates similar samples.
The diversity of generated samples is small compared with those of the original dataset.
As discussed in previous work (e.g. \cite{tran-eccv-2018}), with mode collapse, the generator would map two far-apart latent samples to nearby data points in the high-dimensional data space with high probability. This observation  motivates our idea to constrain  the  distances between generated data points in order to alleviate mode collapse. In particular, the data point distances and the corresponding latent sample distances should be consistent.  

The motivation of our neighbors-embedding constraint $V_R(E, G)$ is to constrain the relative distance among data points and their corresponding latent points within the data and latent manifold respectively (Fig. \ref{neighbor_embedding}). In our model, we apply the probabilistic relative distance (PRDist) in t-SNE \cite{maaten-jmlr-2008}, which takes into account the distributions of  latent sample structure and  data sample structure. 
t-SNE has been shown to  preserve both the local structure of data space (the relation inside each cluster) and the global structure (the relation between each pair of clusters). 
Notably,  our method applies PRDist in the reverse direction of t-SNE for different purpose. While t-SNE aims  to preserve  significant structures of the high-dimensional data  in the reduced-dimensional samples, in our work, we aim  to preserve the structures in low-dimensional latent samples in its high-dimensional mappings via the generator.
Specifically, the objective is as shown in Eq. \ref{eq:kl}:

\begin{equation} 
\mathcal{V}_R(E,G) =  \sum_i \sum_j {p_{i,j} \log {\frac{p_{i,j}}{q_{i,j}}}}
\label{eq:kl}
\end{equation}

The probability distribution of latent structure $p_{i,j}$ is a joint, symmetric distribution, computed as below:

\begin{equation} \label{eu_latent}
p_{i,j} = \frac{p_{i|j} + p_{j|i}}{2n}
\end{equation}
$p_{i|j}$ and $p_{j|i}$ are the conditional probabilities, whose center points are $\mathrm{z}_j$ and $\mathrm{z}_i$ respectively. Here, $i$ and $j$ are indices of $i$-th and $j$-th samples respectively in a mini-batch of training data. Accordingly, $\mathrm{z}_i$ and $\mathrm{z}_j$ are $i$-th and $j$-th latent samples. $n$ is the number of samples in the mini-batch. The conditional probability $p_{j|i}$ is given by:

\begin{equation} \label{p_ij_cond}
p_{j|i} = \frac{{(1 + ||\mathrm{z}_j - \mathrm{z}_i||^2 / 2 \sigma^{2}_{\mathrm{z}})}^{-1}}{\sum_{k\neq i} (1 + ||\mathrm{z}_k - \mathrm{z}_i||^2/2\sigma^{2}_{\mathrm{z}})^{-1}}
\end{equation} 
where $\sigma_\mathrm{z}$ is the variance of all pairwise distances in  a mini-batch of latent samples. Similar to t-SNE method, the joint distribution $p_{i,j}$ is to prevent the problem of outliers in high-dimensional space.

Similarly, the probability distribution of data sample structure $q_{i,j}$ is the joint, symmetric computed from two conditional probabilities as below:

\begin{equation} \label{eu_data}
q_{i,j} = \frac{q_{i|j} + q_{j|i}}{2n}
\end{equation}
where $q_{j|i}$ is the conditional probability of pairwise distance between samples $G(z_j)$ and the center point $G(z_i)$, computed as follow: 

\begin{equation} \label{p_ij_cond}
q_{j|i} = \frac{{(1 + ||G(\mathrm{z}_j) - G(\mathrm{z}_i)||^2 / 2 \sigma^{2}_{\mathrm{x}})}^{-1}}{\sum_{k\neq i} (1 + ||G(\mathrm{z}_k) - G(\mathrm{z}_i)||^2/2\sigma^{2}_{\mathrm{x}})^{-1}}
\end{equation}
$\sigma_\mathrm{x}$ is the variance of all pairwise distances of data samples in the mini-batch. The regularization term $V_R(E,G)$ is the dissimilarity between two joint distributions: $p_{i,j}$ and $q_{i,j}$,  where each distribution represents the neighbor distance distribution.
Similar to t-SNE, we set the values of $p_{i,i}$ and $q_{j,j}$ to zero. The dissimilarity is  Kullback-Leibler (KL) divergence as in Eq. \ref{eq:kl}. $\{\mathrm{z}_i\}$ is  a merged dataset of encoded and random latent samples, and $\{G({\mathrm{z}_i})\}$ is  a merged dataset of reconstruction and generated samples. Here, the reconstruction samples and their latent samples are considered as the anchor points of data and latent manifolds respectively to regularize the generation process.


\subsection{Discriminator objective}

\begin{equation}
\begin{split}
&\mathcal{V}_D(D,G) \\&= (1-\alpha)\mathbb{E}_{\mathrm{x}} \log  D(\mathrm{x}) + \alpha V_C + \mathbb{E}_{\mathrm{z}} \log(1 - D(G(\mathrm{z}))\\
&- \lambda_{\mathrm{p}}V_P
\end{split}
\label{eq_D_log_obj}
\end{equation}

Our discriminator objective is shown in Eq. \ref{eq_D_log_obj}. Our model considers the reconstructed samples as ``real'' represented by the term $V_C = \mathbb{E}_{\mathrm{x}} \log D(G(E(\mathrm{x}))$, so that the gradients from discriminator are not saturated too quickly. 
This constraint slows down the convergence of discriminator, similar goal as   \cite{arjovsky-arxiv-2017},  \cite{miyato-iclr-2018} and \cite{tran-eccv-2018}.
In our method, we use a small  weight for $V_C$ with $\alpha = 0.05$ for the discriminator objective. We observe that $V_C$ is important at the beginning of training. However, towards the end, especially for complex image datasets, the reconstructed samples  may  not be as good as real samples, resulting in  low quality of generated images. Here,  $\mathbb{E}$ is the expectation, $\lambda_{\mathrm{p}}$ is a constant, $V_P = \mathbb{E}_\mathrm{x}(||\nabla_{\hat{\mathrm{x}}} D(\hat{\mathrm{x}})|| - 1)^2$ and $\hat{\mathrm{x}} = \mu \mathrm{x} + (1 - \mu) G(\mathrm{z})$, $\mu$ is a uniform random number $\mu \in U[0,1]$. $V_P$  enforces sufficient gradients  from the discriminator even when approaching convergence. 
Fig. \ref{fig-1d-toy} illustrates gradients at convergence time.

We also apply  hinge loss similar to  \cite{miyato-iclr-2018} by replacing $\log(D(x))$ with $\min(0, -1 + D(x))$. We  empirically found that hinge loss could also improve the quality of generated images in our model. Here, because $D(x) \in (0,1)$, the hinge loss version of Eq. \ref{eq_D_log_obj} (ignore constants) is as follows:

\begin{equation}\label{eq_D_hinge_obj}
\begin{split}
\mathcal{V}^h_D(D,G) = (1 - \alpha)\mathbb{E}_{\mathrm{x}} D(\mathrm{x}) + \alpha V_C - \mathbb{E}_{\mathrm{z}} D(G(\mathrm{z})) - \lambda_{\mathrm{p}}V_P
\end{split}
\end{equation}

\subsection{Generator objective with gradient matching}

In this work, we propose to train the generator via aligning distributions of generated samples and real samples. However, it is challenging to work with high-dimensional sample distribution. We propose to overcome this issue in GAN by using the scalar discriminator scores. In GAN,  the discriminator differentiates real and fake samples. Thus, the discriminator score $D(\mathrm{x})$ can be viewed as the probability that sample $\mathrm{x}$ drawn from the real  data distribution. Although exact form of  $D(\mathrm{x})$ is unknown, but the scores $D(\mathrm{x})$ at some data points $\mathrm{x}$ (from training data) can be computed via the discriminator network. Therefore, we align the distributions by minimizing the difference between discriminator scores of real and generated samples. In addition, we  constrain the gradients of these discriminator scores. These constraints can be derived from Taylor approximation of discriminator functions as followings.

Assume that the first derivative of $D$ exists, and the training set has data samples $\{\mathrm{x}\}$. For a sample point $\mathrm{s}$, by first-order Taylor expansion (TE), we can approximate $D(\mathrm{s})$ with TE at a data point $\mathrm{x}$:

\begin{equation}
\begin{split}
D(\mathrm{s}) &= D(\mathrm{x}) + \nabla_{\mathrm{x}}D(\mathrm{x})(\mathrm{s} - \mathrm{x}) + \epsilon(\mathrm{s}, \mathrm{x})
\end{split}
\label{eq:taylor_x}
\end{equation}
Here $\epsilon(.)$ is the TE approximation error. Alternatively, we can approximate $D(\mathrm{s})$ with TE at a generated sample $G(\mathrm{z})$:

\begin{equation}
\begin{split}
D(\mathrm{s}) &= D(G(\mathrm{z})) + \nabla_{\mathrm{x}}D(G(\mathrm{z}))(\mathrm{s} - G(\mathrm{z})) + \epsilon(\mathrm{s}, G(\mathrm{z}))
\end{split}
\label{eq:taylor_xg}
\end{equation}


Our goal is to enforce the distribution of generated sample $p(G(\mathrm{z}))$ to be similar to that of real sample $p(\mathrm{x})$. 
For a given $\mathrm{s}$, its discriminator score $D(\mathrm{s})$ can be approximated by first-order TE at $\mathrm{x}$ with error $\epsilon(\mathrm{s}, \mathrm{x})$.
Note that, here we define $\epsilon(\mathrm{s}, \mathrm{x})$ to be the approximation error of $D(\mathrm{s})$ with first-order TE at point $\mathrm{x}$. 
Likewise, $\epsilon(\mathrm{s}, G(\mathrm{z}))$ is the approximation error of $D(\mathrm{s})$ with first-order TE at point $G(\mathrm{z})$. 
If $\mathrm{x}$ and $G(\mathrm{z})$ were from the same distribution, then 
 $\mathbb{E}_{\mathrm{x}}\epsilon(\mathrm{s}, \mathrm{x}) \approx \mathbb{E}_{\mathrm{z}}\epsilon(\mathrm{s}, G(\mathrm{z}))$.
Therefore, we propose to enforce $\mathbb{E}_{\mathrm{x}}\epsilon(\mathrm{s}, \mathrm{x}) = \mathbb{E}_{\mathrm{z}}\epsilon(\mathrm{s}, G(\mathrm{z}))$ when training the generator.
Note that $\mathbb{E}_{\mathrm{x}}(D(\mathrm{s})) = \mathbb{E}_{\mathrm{z}}(D(\mathrm{s})) = D(\mathrm{s})$,  because $D(\mathrm{s})$ is a constant and is independent of $\mathrm{x}$ and $\mathrm{z}$.
Therefore, we propose to enforce $\mathbb{E}_{\mathrm{x}}(D(\mathrm{s})) - \mathbb{E}_{\mathrm{x}}\epsilon(\mathrm{s}, \mathrm{x}) = \mathbb{E}_{\mathrm{z}}(D(\mathrm{s})) - \mathbb{E}_{\mathrm{z}}\epsilon(\mathrm{s}, G(\mathrm{z}))$ in order to align $p(G(\mathrm{z}))$ to real sample distribution $p(\mathrm{x})$.
From Eq. \ref{eq:taylor_x}, we have:

\begin{equation}
\begin{split}
&\mathbb{E}_{\mathrm{x}}(D(\mathrm{s})) - \mathbb{E}_{\mathrm{x}}\epsilon(\mathrm{s}, \mathrm{x}) \\
&= \mathbb{E}_{\mathrm{x}}(D(\mathrm{x})) + \mathbb{E}_{\mathrm{x}}(\nabla_{\mathrm{x}}D(\mathrm{x})(\mathrm{s} - \mathrm{x})) \\
& = \mathbb{E}_{\mathrm{x}}(D(\mathrm{x})) + \mathbb{E}_{\mathrm{x}}\nabla_{\mathrm{x}}D(\mathrm{x})\mathrm{s} - \mathbb{E}_{\mathrm{x}}\nabla_{\mathrm{x}}D(\mathrm{x})\mathrm{x}
\end{split}
\label{eq:TE1}
\end{equation} 
From Eq. \ref{eq:taylor_xg}, we have:

\begin{equation}
\begin{split}
& \mathbb{E}_{\mathrm{z}}(D(\mathrm{s})) - \mathbb{E}_{\mathrm{z}}\epsilon(\mathrm{s}, G(\mathrm{z})) \\ 
& = \mathbb{E}_{\mathrm{z}}(D(G(\mathrm{z}))) + \mathbb{E}_{\mathrm{z}}\nabla_{\mathrm{x}}D(G(\mathrm{z}))\mathrm{s} - \mathbb{E}_{\mathrm{z}}\nabla_{\mathrm{x}}D(G(\mathrm{z}))G(\mathrm{z})
\end{split}
\label{eq:TE2}
\end{equation} 
To equate Eqs. \ref{eq:TE1} and \ref{eq:TE2}, we enforce equality of corresponding terms. This leads to minimization of the following objective function for the generator:


\begin{equation}
\begin{split}
& \mathcal{V}_G(D,G) = ||\mathbb{E}_{\mathrm{x}} D(\mathrm{x}) - \mathbb{E}_{\mathrm{z}} D(G(\mathrm{z}))||\\ 
& +  \lambda_{\mathrm{m}}^{1}||\mathbb{E}_{\mathrm{x}}(\nabla_{\mathrm{x}} D(\mathrm{x})) - \mathbb{E}_{z}(\nabla_{\mathrm{x}} D(G(\mathrm{z})))||^2 \\
& + \lambda_{\mathrm{m}}^{2}||\mathbb{E}_{\mathrm{x}}(\nabla_{\mathrm{x}} D(\mathrm{x})^T\mathrm{x}) - \mathbb{E}_{z}(\nabla_{\mathrm{x}} D(G(\mathrm{z}))^TG(\mathrm{z}))||^2 \\
\end{split}
\label{eq:genobj_new}
\end{equation}
Here, we use $\ell_1$-norm for the first term of generator objective, and $\ell_2$-norm for two last terms. Empirically, we observe that using $\ell_2$-norm is more stable than using $\ell_1$-norm. $\lambda_{\mathrm{m}}^{1} = \lambda_{\mathrm{m}}^{2} = 1.0$. In practice, our method is more stable when we implement   $\mathbb{E}_{\mathrm{x}}(\nabla_{\mathrm{x}} D(\mathrm{x}))$ as $\mathbb{E}_{\mathrm{x}}||\nabla_{\mathrm{x}} D(\mathrm{x})||$ and $\mathbb{E}_{\mathrm{x}}(\nabla_{\mathrm{x}} D(\mathrm{x})^T\mathrm{x})$ as $\mathbb{E}_{\mathrm{x}}||\nabla_{\mathrm{x}} D(\mathrm{x})^T\mathrm{x}||$ in the second and third term of Eq. \ref{eq:genobj_new}. 
Note that this  proposed objective can be used in other  GAN models.
Note also that a recent work \cite{tran-eccv-2018} has also used the discriminator score as constraint. However, our motivation and formulation are significantly different. In the experiment, we show improved performance compared to \cite{tran-eccv-2018}.

%% file: experiment.tex
\section{Experimental Results}

\subsection{Synthetic 1D dataset}

\begin{figure}
\centering
\includegraphics[scale=0.2]{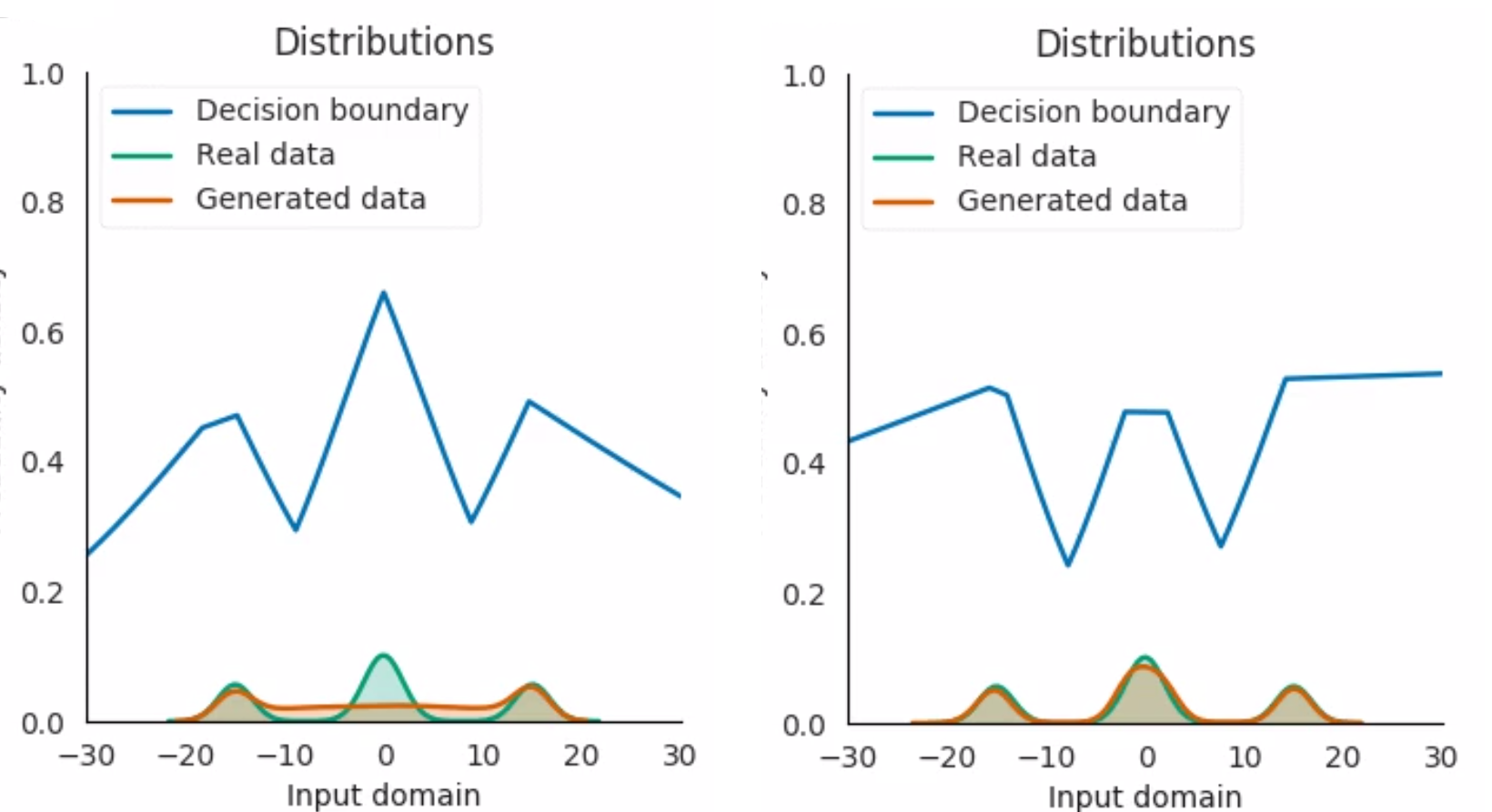}
\caption{We compare our method and Dist-GAN (Tran, Bui and Cheung 2018) on the 1D synthetic dataset of three Gaussian modes. Figures are the last frames of the demo videos (can be found here: https://github.com/tntrung/gan). The blue curve is discriminator scores, the green and orange modes are the training data the generated data respectively.}
\label{fig-1d-toy}
\end{figure}

For 1D synthetic dataset, we compare our model to Dist-GAN \cite{tran-eccv-2018}, a  recent state-of-the-art GAN. We use the code (https://github.com/tntrung/gan) for this 1D experiment. Here, we construct the 1D synthetic data with 3 Gaussian modes (green) as shown in Fig. \ref{fig-1d-toy}. It is more challenging than the one-mode demo by Dist-GAN.

We use small networks for both methods. Specifically, we create the encoder and generator networks with three fully-connected layers and the discriminator network with two fully-connected layers. We use ReLU for hidden layers and sigmoid for the output layer of the discriminator. The discriminator is smaller than the generator to make the training more challenging. The number of neurons for each hidden layer is 4, the learning rate is 0.001, $\lambda_\mathrm{p} = 0.1$ for both method, $\lambda_\mathrm{m}^1 = \lambda_\mathrm{m}^2 = 0.1$ for our generator objective.

Fig. \ref{fig-1d-toy} shows that our model can recover well three modes, while  Dist-GAN cannot (see attached video demos in the supplementary material). Although both methods have good gradients of the discriminator scores (decision boundary) for the middle mode, it's difficult to recover this mode with Dist-GAN as gradients computed over generated samples are not explicitly forced to resemble those of real samples as in our proposed method. Note that for this 1D experiment and the 2D experiment in the next section, we only evaluate our model with gradient matching (+GM), since we find that our new generator with gradient matching alone is already good enough; neighbors embedding is more useful for high-dimensional data samples, as will be discussed.

\subsection{Synthetic 2D dataset}

\begin{figure}
\centering
\includegraphics[scale=0.23]{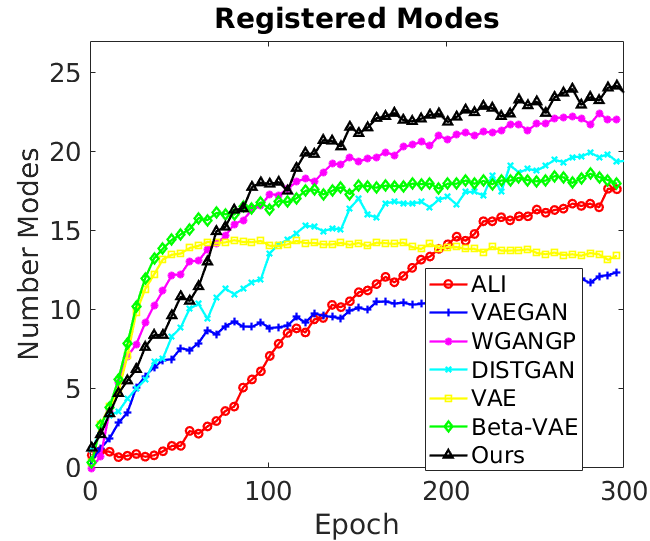}
\includegraphics[scale=0.23]{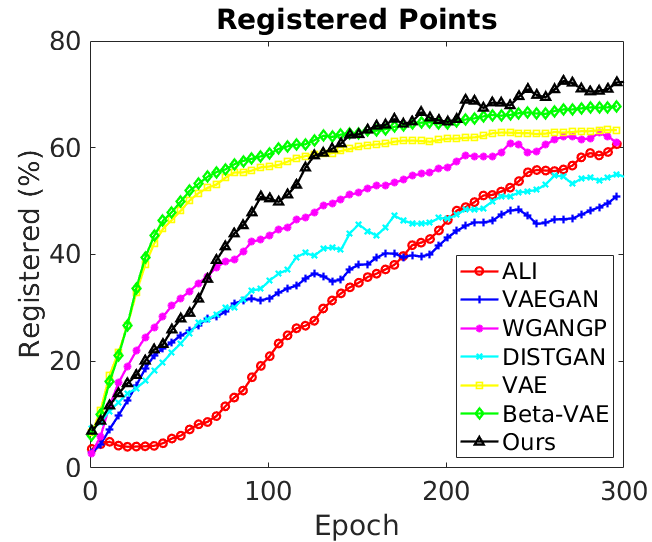}
\caption{Examples of the number of modes (classes) and registered points of compared methods.}
\label{fig-2e-mode-point}
\end{figure}

\begin{figure*}
\centering
\includegraphics[scale=0.52]{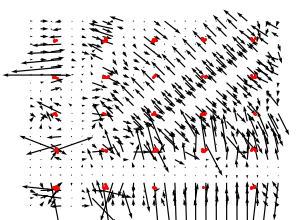}
\includegraphics[scale=0.53]{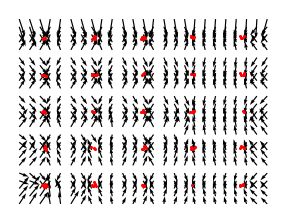}
\includegraphics[scale=0.525]{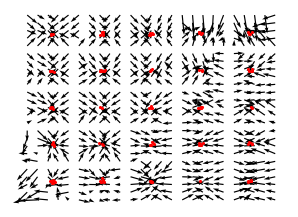}
\includegraphics[scale=0.52]{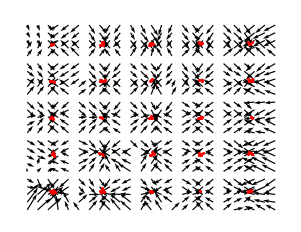}
\caption{Our 2D synthetic data has 25 Gaussian modes (red dots). The black arrows are gradient vectors of the discriminator computed around the ground-truth modes. Figures from left to right are examples of gradient maps of GAN, WGAN-GP, Dist-GAN and ours.}
\label{fig-2d-toy-gradients}
\end{figure*}

For 2D synthetic data, we follow the experimental setup on the same 2D synthetic dataset \cite{tran-eccv-2018}. The dataset has 25 Gaussian modes in the grid layout (red points in Fig. \ref{fig-2d-toy-gradients}) that contains 50K training points. We draw 2K generated samples for evaluating the generator. However, the performance reported in \cite{tran-eccv-2018} is nearly saturated. For example, it can re-cover entirely 25 modes and register more than 90\% of the total number of points. It's hard to see the significant improvement of our method in this case. Therefore, we decrease the number of hidden layers and their number of neurons for networks to be more challenging. For a fair comparison, we use equivalent encoder, generator and discriminator networks for all compared methods.

\begin{table}
\centering
\caption{Network structures for 1D synthetic data in our experiments.}
\begin{tabular}{c | c c c c}
           & $d_\mathrm{in}$ & $d_\mathrm{out}$ & $N_\mathrm{h}$ & $d_\mathrm{h}$ \\ 
\hline
\hline
Encoder ($E$)        & 2     & 2    & 2    & 64 \\
\hline
Generator ($G$)      & 2     & 2    & 2    & 64 \\
\hline
Discriminator ($D$)  & 2     & 1    & 2    & 64 \\
\end{tabular}
\label{toy_network}
\end{table}

The detail of network architecture is presented in Table \ref{toy_network}. $d_\mathrm{in} = 2$, $d_\mathrm{out} = 2$, $d_\mathrm{h} = 64$ are dimensions of input, output and hidden layers respectively. $N_\mathrm{h}$ is the number of hidden layers. We use ReLU for hidden layers and sigmoid for output layers. To have a fair comparison, we carefully fine-tune other methods to ensure that they can perform their best on the synthetic data. For evaluation, a mode is missed if there are less than 20 generated samples registered in this mode, which is measured by its mean and variance of 0.01. A method has mode collapse if there are missing modes. For this experiment, the prior distribution is the 2D uniform $[-1, 1]$. We use Adam optimizer with learning rate lr = 0.001, and the exponent decay rate of first moment $\beta_1 = 0.8$. The parameters of our model are: $\lambda_\mathrm{p} = 0.1, \lambda_\mathrm{m}^1=\lambda_\mathrm{m}^2=0.1$. The learning rate is decayed every 10K steps with a base of $0.99$. This decay rate is to avoid the learning rate saturating too quickly that is not fair for slow convergence methods. The mini-batch size is 128. The training stops after 500 epochs.

In this experiment, we compare our model to several state-of-the-art methods. ALI \cite{donahue-arxiv-2016}, VAE-GAN \cite{larsen-arxiv-2015} and Dist-GAN \cite{tran-eccv-2018} are recent works using
encoder/decoder in their models. WGAN-GP \cite{gulrajani-arxiv-2017} is one of the state-of-the-arts. We also compare to VAE-based methods: VAE \cite{kingma-iclr-2014} and $\beta$-VAE \cite{higgins-iclr-2017}. The numbers of covered (registered) modes and registered points during training are presented in Fig. \ref{fig-2e-mode-point}. The quantitative numbers of last epochs are in Table \ref{tbl_toydata_2d}. In this table, we also report the Total Variation scores to measure the mode balance \cite{tran-eccv-2018}. The result for each method is the average of eight runs. Our method outperforms all others on the number of covered modes. Although WGAN-GP and Dist-GAN are stable with larger networks and this experimental setup \cite{tran-eccv-2018}, they are less stable with our network architecture, miss many modes and sometimes diverge.VAE based method often address well mode collapse, but in our experiment setup where the small networks may affect the reconstruction quality, consequently reduces their performance. Our method does not suffer serious mode collapse issues for all eight runs. Furthermore, we achieve a higher number of registered samples than all others. Our method is also better than the rest with Total Variation (TV).

\begin{table*}
\centering
\footnotesize
\caption{Results on 2D synthetic data. Columns indicate the number of covered modes, and the number of registered samples among 2000 generated samples, and two types of Total Variation (TV). We compare our model to state of the art models: WGAN-GP and Dist-GAN.}
\begin{tabular}{c | c | c | c | c}
\textbf{Method} & \textbf{\#registered modes} & \textbf{\#registered points} & \textbf{TV (True)} & \textbf{TV (Differential)} \\ 
\hline
\hline
GAN \cite{goodfellow-nisp-2014}       & 14.25 $\pm$ 2.49 & 1013.38 $\pm$ 171.73 & 1.00 $\pm$ 0.00 & 0.90 $\pm$ 0.22 \\
\hline
ALI \cite{donahue-arxiv-2016}         & 17.81 $\pm$ 1.80 & 1281.43 $\pm$ 117.84 & 0.99 $\pm$ 0.01 & 0.72 $\pm$ 0.19 \\
\hline
VAEGAN \cite{larsen-arxiv-2015}       & 12.75 $\pm$ 3.20 & 1042.38 $\pm$ 170.17 & 1.35 $\pm$ 0.70 & 1.34 $\pm$ 0.98 \\
\hline
VAE \cite{kingma-iclr-2014}  & 13.48 $\pm$ 2.31 & 1265.45 $\pm$ 72.47 & 1.81 $\pm$ 0.71 & 2.16 $\pm$ 0.72 \\
\hline
$\beta$-VAE \cite{higgins-iclr-2017}   & 18.00 $\pm$ 2.33 & 1321.17 $\pm$ 95.61 & 1.17 $\pm$ 0.24 & 1.47 $\pm$ 0.28 \\
\hline
WGAN-GP \cite{gulrajani-arxiv-2017}   & 21.71 $\pm$ 1.35 & 1180.25 $\pm$ 158.63 & 0.90 $\pm$ 0.07 & 0.51 $\pm$ 0.06 \\
\hline
Dist-GAN \cite{tran-eccv-2018}		  & 20.71 $\pm$ 4.42 & 1188.62 $\pm$ 311.91 & 0.82 $\pm$ 0.19 & 0.43 $\pm$ 0.12 \\
\hline
\textbf{Ours}   				      & 24.39 $\pm$ 0.44 & 1461.83 $\pm$ 222.86 & 0.57 $\pm$ 0.17 & 0.31 $\pm$ 0.12 \\
\end{tabular}
\label{tbl_toydata_2d}
\end{table*}

In addition, we follow \cite{tung-icmlw-2018} to explore the gradient map of the discriminator scores of compared methods: standard GAN, WGAN-GP, Dist-GAN and ours as shown in Fig. \ref{fig-2d-toy-gradients}. This map is important because it shows the potential gradient to pull the generated samples towards the real samples (red points). The gradient map of standard GAN is noisy, uncontrolled and vanished for many regions. The gradient map of WGAN-GP has more meaningful directions than GAN. Its gradient concentrates in the centroids (red points) of training data and has gradients around most of the centroids. However, WGAN-GP still has some modes where  gradients are not towards the ground-truth centroids. Both Dist-GAN and our method show better gradients than WGAN-GP. The gradients of our method are more informative for the generator to learn when they guide better directions to all real ground-truths.

\subsection{CIFAR-10 and STL-10 datasets}

\begin{figure}
\centering
\includegraphics[scale=0.60]{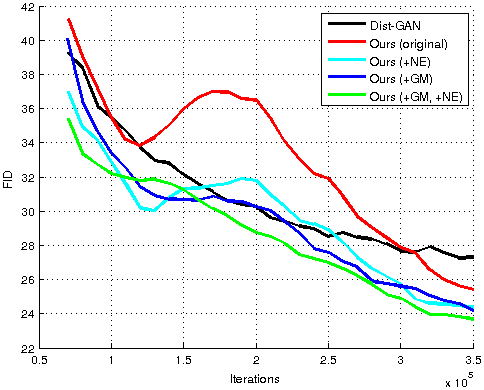}
\caption{FID scores of our method compared to Dist-GAN.}
\label{fig:lg_obj_fid}
\end{figure}

For CIFAR-10 and STL-10 datasets, we measure the performance with FID scores \cite{heusel-arxiv-2017}. FID can detect intra-class mode dropping, and measure the diversity as well as the quality of generated samples. We follow the experimental procedure and model architecture in \cite{miyato-iclr-2018} to compare methods. FID is computed from 10K real samples and 5K generated samples. Our default parameters are used for all experiments $\lambda_\mathrm{p} = 1.0, \lambda_\mathrm{r} = 1.0, \lambda^1_\mathrm{m} = \lambda^2_\mathrm{m} = 1.0$. Learning rate, $\beta_1$, $\beta_2$ for Adam is (lr = 0.0002, $\beta_1 = 0.5$, $\beta_2= 0.9$). The generator is trained with 350K updates for logarithm loss version (Eq. \ref{eq_D_log_obj}) and 200K for ``hinge" loss version (Eq. \ref{eq_D_hinge_obj}) to converge better. The dimension of the prior input is 128. All our experiments are conducted using the unsupervised setting.

\begin{table}
\centering
\footnotesize
\caption{Comparing the FID score to the state of the art (Smaller is better). Methods with the CNN and ResNet (R) architectures. FID scores of SN-GAN, Dist-GAN and our method reported with hinge loss. Results of compared methods are from \cite{miyato-iclr-2018,tran-eccv-2018}.}
\begin{tabular}{ c | l | c | l}
        \textbf{Method} & \textbf{CIFAR} & \textbf{STL} & \textbf{CIFAR (R)}\\
\hline
\hline
GAN-GP     			         	 & 37.7   & -     & -                    \\
WGAN-GP    			        	 & 40.2   & 55.1  & -                    \\
SN-GAN     			        	 & 25.5   & 43.2  & 21.7 $\pm$ .21       \\
Dist-GAN   			        	 & 22.95  & 36.19 & -                    \\
\textbf{Ours}                    & 21.70  & 30.80 & 16.47 $\pm$ .28      \\    
\end{tabular}
\label{fid_score}
\end{table}

\begin{figure*}
\centering
\includegraphics[scale=0.50]{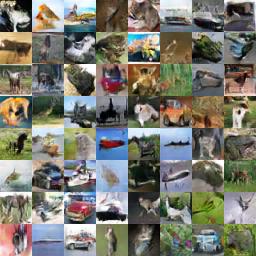}
\includegraphics[scale=0.50]{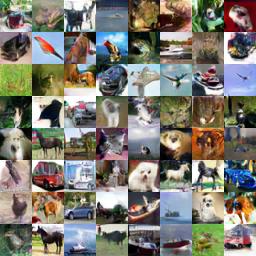}
\includegraphics[scale=0.335]{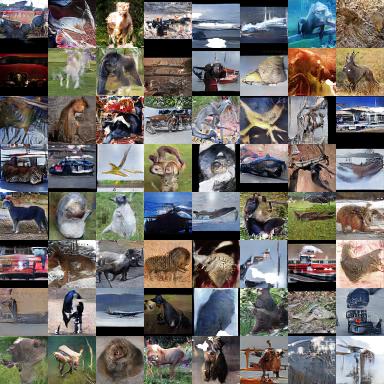}
\caption{Generated samples of our method. Two first samples are on CIFAR-10 with CNN and ResNet architectures, and the last one is on STL-10 with CNN.}
\label{fig:examples}
\end{figure*}

In the first experiment, we conduct the ablation study with our new proposed techniques to understand the contribution of each component into the model. Experiments with standard CNN \cite{miyato-iclr-2018} on the CIFAR-10 dataset. We use the logarithm version for the discriminator objective (Eq. \ref{eq_D_log_obj}). Our original model is similar to Dist-GAN model, but we have some modifications, such as: using lower weights for the reconstruction constraint as we find that it can improve FID scores. We consider Dist-GAN as a baseline for this comparison. FID is computed for every 10K iterations and shown in Fig. \ref{fig:lg_obj_fid}. Our original model converges a little bit slow at the beginning, but at the end, our model achieves better FID score than Dist-GAN model. Once we replace separately each proposed techniques, either the neighbors embedding technique (+NE) or gradient matching (+GM), into our original model, it converges faster and reaches a better FID score than the original one. Combining two proposed techniques further speeds up the convergence and reach better FID score than other versions. This experiment proves that our proposed techniques can improve the diversity of generated samples. Note that in Fig. \ref{fig:lg_obj_fid}, we compared Dist-GAN and our model (original) with only discriminator scores. With GM, our model converges faster and achieves better FID scores.



We compare our best setting (NE + GM) with a hinge loss version (Eq. \ref{eq_D_hinge_obj}) with other methods. Results are shown in Table \ref{fid_score}. The FID score of SN-GAN and Dist-GAN are also with hinge loss function. We also report our performance with the ResNet (R) architecture \cite{miyato-iclr-2018} for CIFAR-10 dataset. For both standard CNN and ResNet architectures, our model outperforms other state-of-the-art methods with FID score, especially significantly higher on STL-10 dataset with CNN and on CIFAR-10 dataset with ResNet. For STL-10 dataset and the ResNet architecture, the generator is trained with 200K iterations to reduce training time. Training it longer does not significantly improve the FID score. Fig. \ref{fig:examples} are some generated samples of our method trained on CIFAR-10 and STL-10 datasets.

\begin{figure*}
\centering
\includegraphics[scale=0.60]{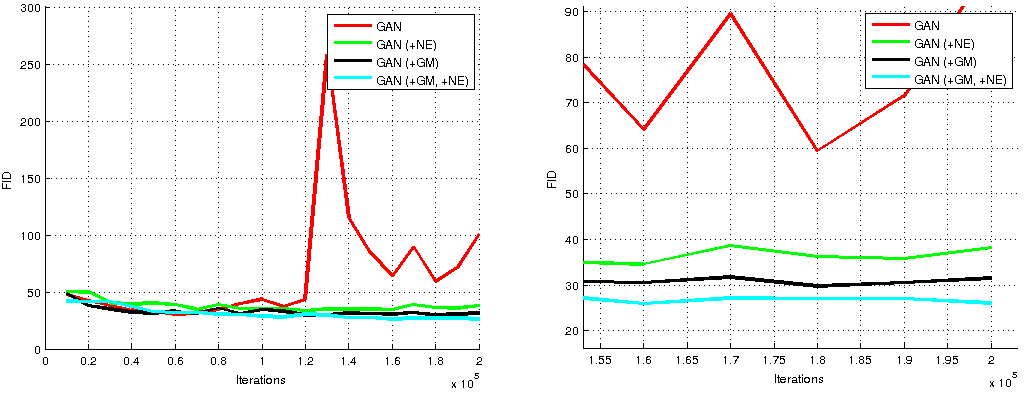}
\caption{FID scores of GAN when applying our proposed techniques for the generator, and its zoomed figure on the right.}
\label{fig:gan_lg_obj_fid}
\end{figure*}

Our proposed techniques are not only usable in our model, but can be used  for other GAN models. We demonstrate this by applying them for standard GAN \cite{goodfellow-nisp-2014}. This experiment is conducted on the CIFAR-10 dataset using the same CNN architecture as \cite{miyato-iclr-2018}. First, we regularize the generator of GAN by our propose neighbors embedding or gradient matching separately or their combination to replace the original generator objective of GAN. When applying NE and GM separately, each of them itself can significantly improve FID as shown in Fig. 6. In addition, from Fig. \ref{fig:gan_lg_obj_fid}, GM+NE achieves FID of 26.05 (last iteration), and this is significant improvement compared to GM alone with FID of 31.50 and NE alone with FID of 38.10. It's interesting that GM can also reduce mode collapse, we let the further investigation of it in the future work. Although both can handle the mode collapse, NE and GM are very different ideas: NE is a manifold learning based regularization to explicitly prevent mode collapse; GM aligns distributions of generated samples and real samples. The results (Figs. \ref{fig:lg_obj_fid} and \ref{fig:gan_lg_obj_fid}) show GM+NE leads to better convergence and FID scores than individual techniques.

To examine the computational time of gradient matching of our proposed generator objective, we measure its training time for one mini-batch (size 64) with/without GM (Computer: Intel Xeon Octa-core CPU E5-1260 3.7GHz, 64GB RAM, GPU Nvidia 1080Ti) with CNN for CIFAR-10. It takes about 53ms and 43ms to train generator for one mini-batch with/without the GM term respectively. For 300K iterations (one mini-batch per iteration), training with GM takes about one more hour compared to without GM. The difference is not serious. Note that GM includes $\ell_1$, $\ell_2$ norms of the difference of discriminator scores and gradients, which can be computed easily in Tensorflow.



%% file: conclusion.tex
\section{Conclusion}

We propose two new techniques to address  mode collapse  and improve the diversity of generated samples. First, 
we propose an inverse t-SNE regularizer to explicitly retain 
local structures of latent samples in the generated samples to reduce mode collapse.
Second, we propose a new gradient matching regularization for the generator objective, which improves convergence and the quality of generated images. We derived this gradient matching constraint from Taylor expansion. Extensive experiments demonstrate that both constraints can improve GAN. The combination of our proposed techniques leads to state of the art FID scores on benchmark datasets. 
Future work applies our model for other applications, such as: person re-identification \cite{yiluan-cvpr-2018}, anomaly detection \cite{lim-icdm-2018}.

%% file: acknowledge.tex
\section*{Acknowledgement}

This work was supported by both ST Electronics and the National Research Foundation(NRF), Prime Minister's Office, Singapore under Corporate Laboratory @ University Scheme (Programme Title: STEE Infosec - SUTD Corporate Laboratory).